\documentclass[letterpaper, 10 pt, conference]{ieeeconf}  

\IEEEoverridecommandlockouts                              

\usepackage{stfloats} 

\usepackage{graphics} 
\usepackage{epsfig} 
\usepackage{mathptmx} 
\usepackage{times} 
\usepackage{amsmath} 
\usepackage{amssymb}  

\usepackage{hyperref}
\usepackage{algorithm}
\usepackage{algpseudocode}
\usepackage{multirow}
\usepackage[table]{xcolor}
\usepackage{lipsum}
\usepackage{comment} 
\usepackage{pifont}
\usepackage{listings}
\usepackage{bm}
\usepackage{footmisc}
\usepackage{balance}
\usepackage{subcaption}  
\usepackage{caption}

\title{\LARGE \bf
Preference Aligned Visuomotor Diffusion Policies \\ for Deformable Object Manipulation
}

\author{Marco Moletta$^1$, Michael C. Welle$^{1,2}$,  Danica Kragic$^1$
\thanks{1-KTH Royal Institute of Technology, Sweden, {\it\small \{moletta, mwelle, dani\}@kth.se}}
\thanks{2-INCAR Robotics AB, Sweden, {\it\small michael.welle@incar-robotics.se}  }
\thanks{This work was supported by the European Union’s Horizon Europe Framework Programme under grant agreement No. 101070596 (euROBIN).}
}

\begin{document}

\maketitle

\thispagestyle{empty}
\pagestyle{empty}

\begin{abstract}
Humans naturally develop preferences for how manipulation tasks should be performed, which are often subtle, personal, and difficult to articulate. Although it is important for robots to account for these preferences to increase personalization and user satisfaction, they remain largely underexplored in robotic manipulation, particularly in the context of deformable objects like garments and fabrics. In this work, we study how to adapt pretrained visuomotor diffusion policies to reflect preferred behaviors using limited demonstrations. We introduce RKO, a novel preference-alignment method that combines the benefits of two recent frameworks: RPO and KTO. We evaluate RKO against common preference learning frameworks, including these two, as well as a baseline vanilla diffusion policy, on real-world cloth-folding tasks spanning multiple garments and preference settings. We show that preference-aligned policies (particularly RKO) achieve superior performance and sample efficiency compared to standard diffusion policy fine-tuning. These results highlight the importance and feasibility of structured preference learning for scaling personalized robot behavior in complex deformable object manipulation tasks.
\end{abstract}

\section{Introduction}

\begin{figure}[t] 
    \centering 
    \includegraphics[width=0.83\linewidth]{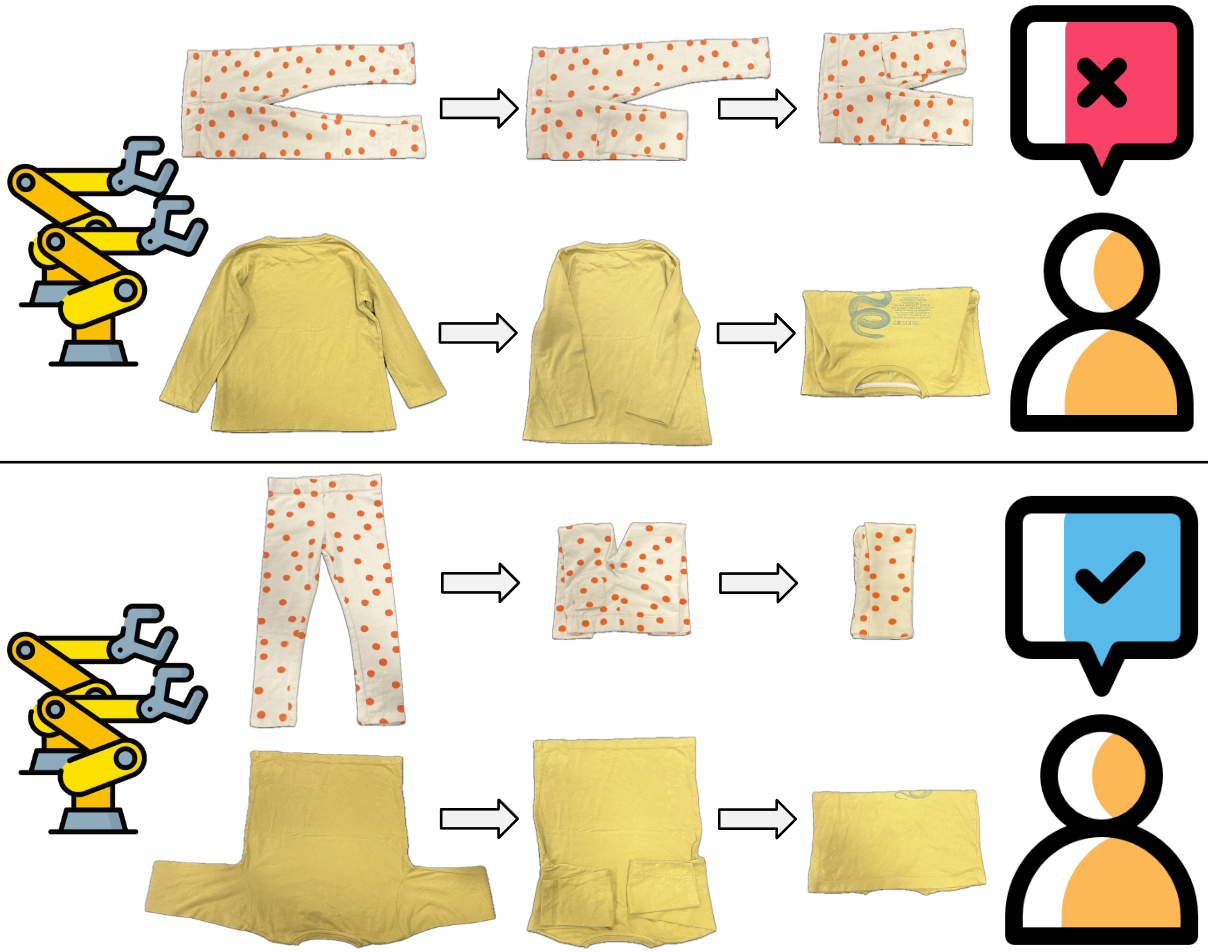} \caption{Two different user preferences for folding the same garments demonstrate how variations in execution can reflect personal styles or practical needs. Capturing and aligning with such preferences is essential for enabling robots to perform personalized and user-aligned behaviors in deformable object manipulation tasks like cloth folding.} 
    \label{fig:first_page} 
    \vspace{-15pt}
\end{figure}

As robots become more affordable and integrated into daily life, there is a growing need for adaptive behaviors that reflect individual user preferences~\cite{canal2019adapting}. Personalization often requires robots to learn directly from interaction~\cite{woodworth2018preference}, with users providing demonstrations of desired behaviors. In this context, a particularly relevant yet underexplored domain is Deformable Object Manipulation (DOM), involving everyday items like clothing, textiles, and food. Enabling robots to handle such objects with human-like dexterity while following human preferences would support applications in laundry folding, assisted dressing, feeding, and healthcare~\cite{zhu2022challenges}. Personalization in DOM may involve different folding styles, dressing strategies tailored to mobility constraints, or household routines reflecting different preferences between users, as illustrated in Fig.\ref{fig:first_page}.

Deformable objects pose unique challenges compared to rigid ones due to their complex dynamics, high-dimensional state spaces, and varied physical properties~\cite{longhini2024unfolding}. These complexities make both manipulation and the expression of user preferences more difficult. While strategies for rigid objects can often be described verbally, the subtleties of tasks like garment folding are harder to articulate with words, making physical demonstrations a more natural modality for expressing preferences~\cite{canal2020adapting}. However, collecting such demonstrations is costly and time-consuming~\cite{moletta2023virtual}, highlighting the need for sample-efficient adaptation frameworks.
Recent advances in data-driven learning have shown great progress in manipulation, including DOM, with robotic foundation models~\cite{black2024pi0} and visuomotor diffusion policies~\cite{chi2023diffusion} leveraging large-scale demonstration datasets to generalize across tasks. However, adapting these pretrained models to reflect user-specific manipulation preferences has received scarce attention. The challenge lies in aligning them to new behaviors without forgetting prior knowledge or requiring many new demonstrations, a common scenario where existing task demonstrations are available and could be leveraged to support the adaptation process to a new preferred behavior.

In this work, we address the problem of accounting for user preferences in deformable object manipulation. Specifically, we study how to align a pretrained visuomotor diffusion policy to new demonstrations reflecting a user’s preferred garment-folding strategy. While preference optimization techniques have seen success in domains like text-to-image generation~\cite{liu2025survey}, their application to robotic DOM remains limited. We investigate direct preference optimization (DPO)~\cite{rafailov2023direct}, along with two recent variants: relative preference optimization (RPO)~\cite{gu2024diffusion}, which leverages observation similarity for contrastive weighting, and Kahneman-Tversky Optimization (KTO)~\cite{li2024aligning}, which enables training from per-sample binary feedback instead of preference pairs. Building on these, we propose RKO, a novel method combining the strengths of KTO and RPO.

We evaluate RKO against standard preference optimization frameworks and vanilla diffusion policies. Results show that preference-based methods, particularly RKO, achieve better alignment to user-preferred behaviors with fewer demonstrations, highlighting their sample efficiency and ability to retain knowledge from pretrained models.

The contributions of this paper are as follows:
\begin{itemize}
\item We introduce RKO, a novel method that combines the sample efficiency of KTO with the context-aware weighting of RPO.
\item We present the first systematic comparison of preference optimization frameworks (DPO, RPO, KTO) for aligning pretrained diffusion policies to user-specific strategies in deformable object manipulation.
\item We conduct extensive real-world evaluations on three garment types (trousers, sleeves, tshirt), each with multiple folding preference tasks, demonstrating the effectiveness and sample efficiency of preference-based alignment over vanilla diffusion policies.
\end{itemize}

\section{Related Work}

We structure the related work along the areas of deformable object manipulation, visuomotor diffusion policies, as well as preference alignment frameworks and their applications in robotics.

\subsection{Learning Deformable Object Manipulation Skills}

Deformable object manipulation (DOM) poses unique challenges due to the high-dimensional, nonlinear dynamics of garments, ropes, and tissues~\cite{laezza2024robot}. Their virtually infinite degrees of freedom complicate perception and state estimation, while analytical deformation models are often computationally prohibitive~\cite{longhini2025adapting}. Data-driven methods have therefore become increasingly prominent. Learning from demonstrations offers a scalable way to acquire complex skills without explicit physics-based modeling~\cite{palan2019learning, ingelhag2024robotic, dalal2023imitating}. In this context, diffusion models~\cite{ho2020denoising} have shown strong ability to capture multimodal distributions across robotics tasks, including navigation~\cite{sridhar2024nomad}, planning~\cite{kapelyukh2023dallebot}, and both rigid and deformable manipulation~\cite{mishra2023generative, reuss2023goal}. Visuomotor diffusion policies learn a stochastic transport map from a simple prior (e.g., Gaussian noise) to a target distribution of action sequences conditioned on observations, enabling better generalization than traditional discriminative models~\cite{chi2023diffusion, pearce2023imitating}. Their ability to capture structured, high-dimensional behaviors directly from demonstrations makes them well suited for DOM, where dynamics are complex and difficult to model explicitly. However, many real-world DOM applications (such as folding clothes, dressing patients, or household routines) are shaped by user-specific preferences, but remain understudied~\cite{canal2020adapting}. Addressing this requires sample-efficient alignment of pretrained policies to diverse demonstrations, which motivates the present work.

\subsection{Preference Alignment in Robotics}

Adapting robot behavior to user-specific needs is often studied through preference learning~\cite{furnkranz2010preference, sadigh2017active}, where predictive models of preferences are inferred from human feedback, typically in the form of pairwise comparisons between task executions (e.g., a user indicating which of two executions they prefer)~\cite{wirth2017survey}. A common approach is reinforcement learning from human feedback (RLHF)~\cite{ cheng2011preference, wilson2012bayesian}, which first trains a reward model on preference data and then fine-tunes a policy to maximize this learned reward~\cite{yin2024relative, woodworth2018preference, wirth2017survey}. RLHF has been pivotal in aligning large language models with human intent, but its dependence on reward modeling and RL optimization limits scalability in robotics~\cite{ouyang2022training, winata2025preference}.

To overcome the limitations of RLHF, direct preference optimization (DPO)~\cite{rafailov2023direct} and related methods bypass reward modeling by directly contrasting preferred and dispreferred behaviors, though they still require paired feedback. Relative preference optimization (RPO)~\cite{yin2024relative} extends this idea by exploiting similarities across demonstrations, weighting all win–lose pairs within a batch to improve alignment. Kahneman-Tversky Optimization (KTO)~\cite{li2024aligning} further reduces data requirements by using per-sample binary labels instead of pairwise comparisons. Diffusion-based preference optimization has shown strong results in text-to-image alignment~\cite{wallace2024diffusion, gu2024diffusion, li2024aligning}, demonstrating scalability to high-dimensional generative models. However, its role in robotic manipulation (especially DOM) remains limited. Prior work has largely addressed rigid-object tasks such as rearrangement with language-based preferences~\cite{lee2024visual}, or narrow DOM scenarios like assisted dressing for improved comfort and safety~\cite{zhang2017personalized, canal2019adapting}, but without leveraging the capabilities of diffusion models. Other studies have examined preference learning in handovers or coactive settings~\cite{cakmak2011human, jain2015learning}, but these remain sparse and often limited to simple tasks.
More recently, diffusion policies combined with DPO have been used to improve data efficiency in long-horizon planning~\cite{yuan2024preference} and even in DOM~\cite{chen2024deformpam}, though not for preference alignment. Bridging this gap by systematically comparing DPO, RPO, and KTO with diffusion-based visuomotor policies for DOM is one of the central focuses of this work.

\begin{figure*}[t!]
  \centering
         \includegraphics[width=0.83\textwidth]{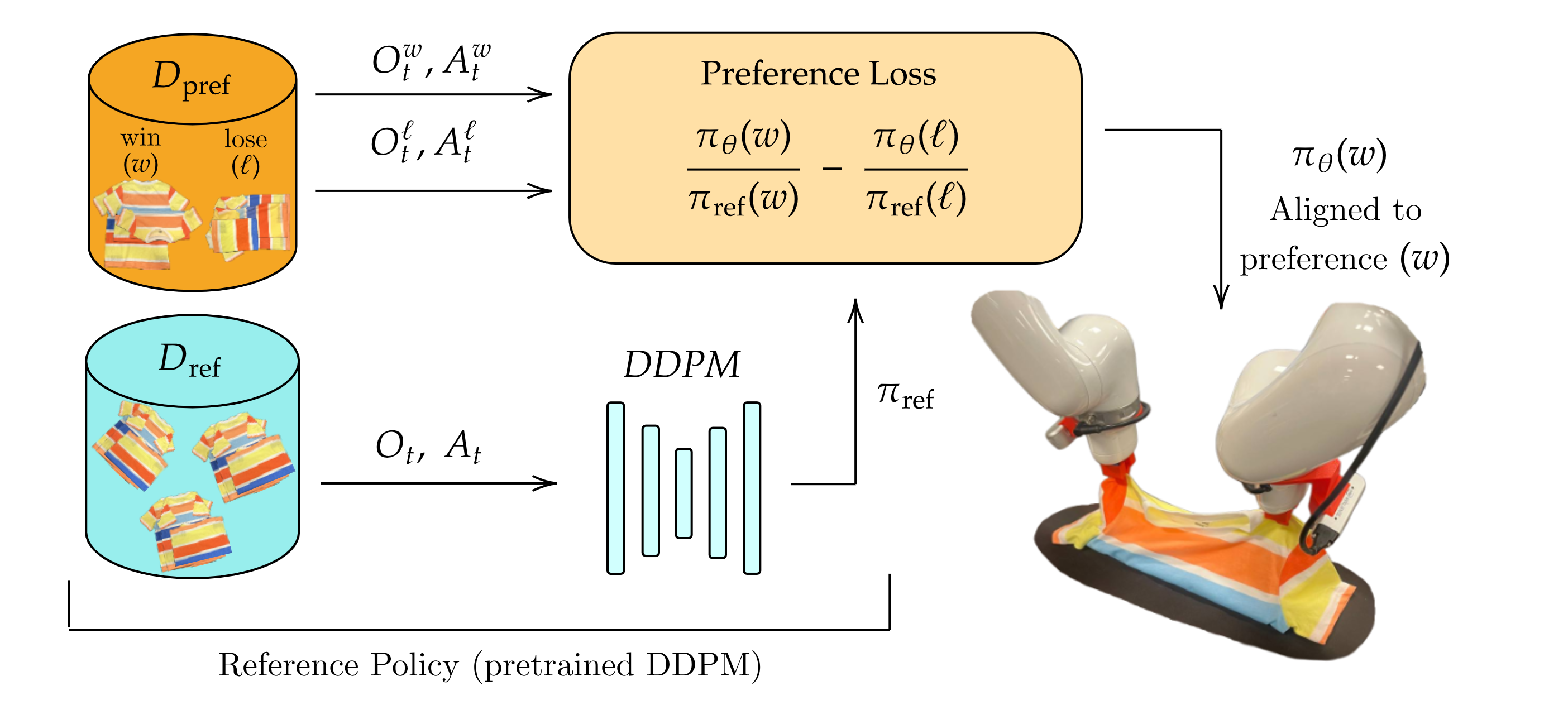}
    \caption{General preference alignment framework used in this work. A reference model is first trained on a large set of demonstrations ($D_{\text{ref}}$) for a given task. To align it to a user’s preferred strategy, a new set of \textit{winning} demonstrations is collected and combined into $D_{\text{pref}}$ along with \textit{losing} demonstrations, i.e., examples of alternative, dispreferred behaviors (which may also come from $D_{\text{ref}}$). The preference loss then aligns the new policy $\pi_{\theta}$ to the preferred behavior by explicitly contrasting it with the \textit{losing} demonstrations. This enables more effective and sample-efficient alignment than training a diffusion policy solely on the \textit{winning} demonstrations.}
    \label{fig:high-level-summary}
    \vspace{-\baselineskip}
\end{figure*}

\section{Method} \label{method}

We begin with a brief overview of diffusion models and the preference alignment frameworks used in this work, followed by our proposed optimization method (RKO) and details on system design.

\subsection{Preliminaries: Visuomotor diffusion models}

The DDPM formulation defines a forward diffusion process (noise addition) and a backward denoising process. Starting from a noise sample $x_K$ drawn from a Gaussian, the model iteratively denoises through $K$ steps toward $x_0=\varepsilon_\theta(x_k,k)$ (a data sample), using a learned noise-prediction (score) network $\varepsilon_\theta$.
For visuomotor control, the formulation is adapted in~\cite{ho2020denoising} with two key modifications: (i) the output $x$ becomes an action sequence, so diffusion operates in action space $A_t$; (ii) the score network is conditioned on observations $O_t$ (e.g., recent visual inputs), modeling the conditional distribution $p(A_t \mid O_t)$. Observation features are extracted once per time step and reused across denoising steps for efficiency and stability.

The denoising step is:
\begin{equation}
A^{k-1}_t = \alpha_k\left( A^k_t - \gamma_k\,\varepsilon_\theta(O_t, A^k_t, k)\right) + \mathcal{N}(0,\sigma_k^2 I),
\end{equation}
with $\alpha_k, \gamma_k, \sigma_k$ forming the noise schedule. Training samples real actions $A_0$, adds noise at a random step $k$, and trains $\varepsilon_\theta$ to predict $\epsilon_k$ via an MSE loss, corresponding to minimizing a variational lower bound:

\begin{equation}
\mathcal{L}_{DDPM}(\theta) = MSE\big(\epsilon_k, \epsilon_{\theta}(O_t, A^0_t + \epsilon_k, k) \big)
\end{equation}

\subsection{Preference Alignment frameworks}

In this section, we detail how a pretrained diffusion model can be aligned to a new preferred behavior, expressed through novel preferred demonstrations, using diffusion-based preference alignment methods (DPO, RPO, KTO, RKO (ours)), highlighting their key similarities and differences. We first outline the common problem setting and how these frameworks are adapted to visuomotor policies.

\textbf{Problem setting:} for each preference optimization framework, we consider a dataset $D_{\text{pref}} = {((O^w, A^w), (O^{\ell}, A^{\ell}))}$, where each example consists of preferred (\textit{win}) demonstrations $(O^w, A^w)$ and dispreferred (\textit{lose}) ones $(O^{\ell}, A^{\ell})$, with labels $w=1, \ell=-1$. Alongside this, we use a pretrained reference diffusion model $\pi_{ref}$, trained on a larger dataset $D_{\text{ref}}$ of preference-free demonstrations. These reference demonstrations represent a default or neutral strategy for performing the task (e.g., folding the same garment in a different but acceptable way). The reference model serves both as a reference for evaluating relative output quality and as initialization for the preference-aligned policy $\pi_{\theta}$, which is then fine-tuned using the preference losses of each framework. In contrast, the lose demonstrations in $D_{\text{pref}}$ are not neutral but explicitly represent a behavior the user does not want (e.g., an undesired folding style) and are included to teach the model to avoid that behavior while learning from the \textit{winning} demonstrations. The goal is to align $\pi_{\theta}$ with the \textit{win} behaviour while using the same \textit{winning} demonstrations across frameworks, enabling a fair comparison of sample efficiency. The preference loss amplifies the distinction between \textit{winning} and \textit{losing} behaviors, guiding the policy towards the preferred strategy while discouraging the dispreferred one. For clarity, we omit the denoising step $k$ and denote $\ell$ for \textit{lose} and $w$ for \textit{win}. The overall pipeline is illustrated in Fig.~\ref{fig:high-level-summary}. Preference-free datasets $D_{\text{ref}}$ are first used to train reference diffusion policies conditioned on observation–action sequences $(O_t, A_t)$. Each reference policy $\pi_{\text{ref}}$ is then used in combination with a novel preference dataset $D_{\text{pref}}$ for preference alignment. For evaluation, we compare the policies trained with our proposed framework ($\pi_{\text{RKO}}$) against the main classes of preference-optimization frameworks ($\pi_{\text{DPO}}, \pi_{\text{RPO}}, \pi_{\text{KTO}}$) and a vanilla DDPM trained only on preference demonstrations ($\pi_{\text{DDPM}}$).

\begin{figure*}[t!]
  \centering
         \includegraphics[width=0.97\textwidth]{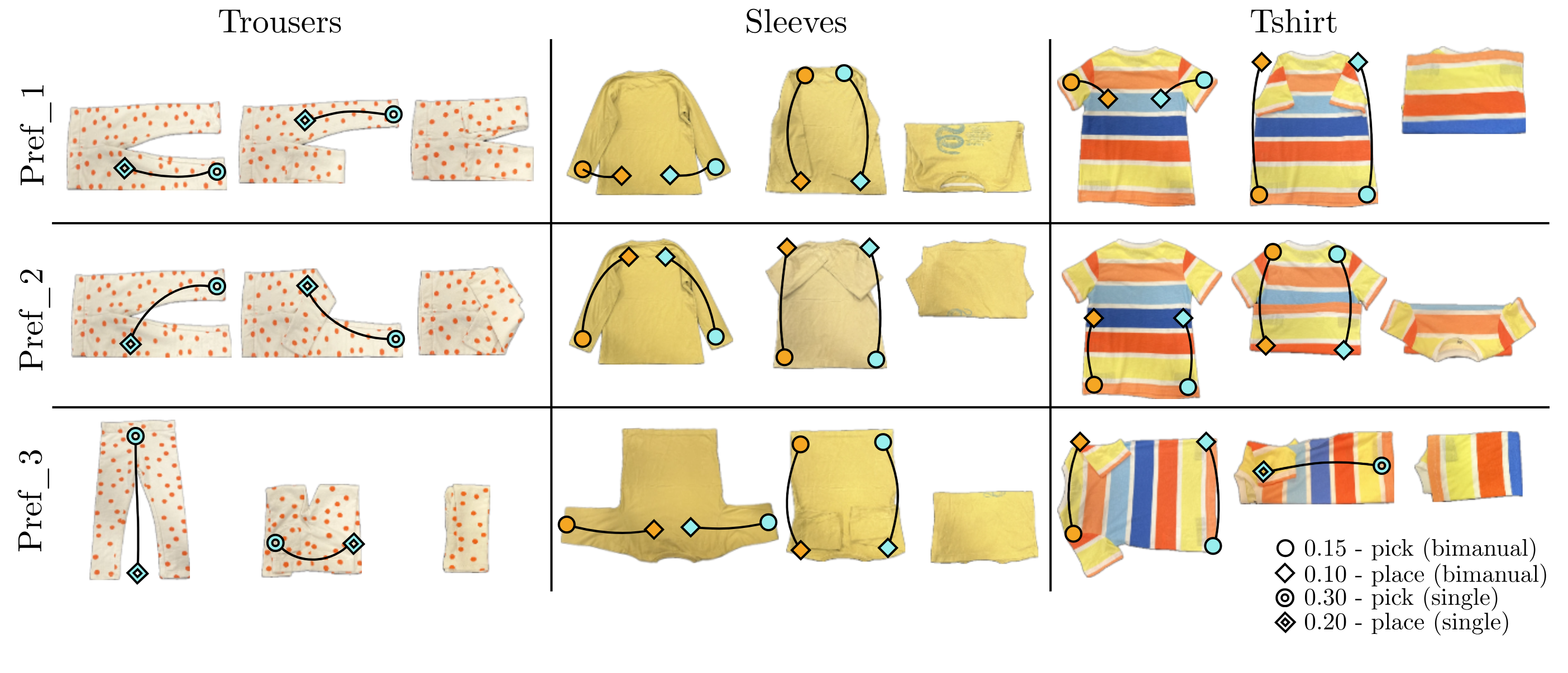}
    \caption{Illustration of three garment-folding preferences for each garment type. Each panel shows the pick (circle) and place (diamond) positions for the left (orange) and right (light blue) arms. The scores are visible on the bottom right: bimanual actions are executed synchronously, and their scores are normalized so that each individual action contributes equally. The total score for a complete, correct folding sequence is $1$.}
    \label{fig:tasks}
    \vspace{-\baselineskip}
\end{figure*}

\subsubsection{Diffusion-DPO}

Unlike RLHF, Diffusion-DPO~\cite{wallace2024diffusion} directly aligns a diffusion model with human preferences, avoiding both reward modeling and reinforcement learning. The reference model serves as a baseline policy, anchoring what is considered “default” behavior and enabling relative comparisons between \textit{win} and \textit{lose} samples.

The objective is defined as the following pairwise loss:
\begin{multline}
\mathcal{L}_{\text{Diffusion-DPO}}(\theta) = \mathbb{E}_{(A_0^w, A_0^\ell) \sim D_{\text{pref}}, \, 
t \sim \mathcal{U}(0, T), \epsilon \sim \mathcal{N}(0, I)}  \\
\quad  \beta \log \sigma \big[ \big(\| \epsilon - \epsilon_\theta(O_t^w, A_t^w, t) \|_2^2 
    - \| \epsilon - \epsilon_{\text{ref}}(O_t^w, A_t^w, t) \|_2^2 \big)\\
\qquad - \big( \| \epsilon - \epsilon_\theta(O_t^\ell, A_t^\ell, t) \|_2^2 
    - \| \epsilon - \epsilon_{\text{ref}}(O_t,^\ell, A_t^\ell, t) \|_2^2 \big) \big] 
\label{eq:dpo}
\end{multline}

where $A_t^w \sim q(A_t^w \mid A_0^w)$ and $A_t^\ell \sim q(A_t^\ell \mid A_0^\ell)$ denote noisy samples at step $t$ obtained from the forward diffusion process, $\epsilon \sim \mathcal{N}(0, I)$ is sampled noise, and $\beta$ is a scaling factor.

This loss measures the relative advantage of the current noise prediction network $\epsilon_\theta$ over the reference one $\epsilon_{\text{ref}}$ when comparing preferred and dispreferred demonstrations.

\subsubsection{Diffusion-RPO}

While Diffusion-DPO relies on single win–lose pairs, it does not exploit the similarity and structure shared across demonstrations. Diffusion-RPO~\cite{yin2024relative} extends this formulation with a context-aware reweighting scheme: each winning sample is compared against all losers in the batch, with importance weights assigned based on semantic similarity.
The loss is defined over all pairs in the batch:
\begin{multline}
\mathcal{L}_{\text{Diffusion-RPO}}(\theta) =
\mathbb{E}_{ (A^w_0, A^\ell_0) \sim D_{\text{pref}}, t \sim \mathcal{U}(0, T), \epsilon \sim \mathcal{N}(0, I)} \\
\;\; \omega_{i,j} \cdot \beta \log \sigma \big[ \big(
\| \epsilon - \epsilon_\theta(O_{t,i}^w, A^w_{t,i}, t) \|_2^2
- \| \epsilon - \epsilon_{\text{ref}}(O_{t,i}^w,A^w_{t,i}, t) \|_2^2\big) \\
-\big( \| \epsilon - \epsilon_\theta(O_{t,j}^\ell, A^\ell_{t,j}, t) \|_2^2
- \| \epsilon - \epsilon_{\text{ref}}(O_{t,j}^\ell, A^\ell_{t,j}, t) \|_2^2
\big)
\big]
\label{eq:rpo}
\end{multline}

where $i$ indexes winning samples and $j$ indexes losing samples within the batch. The key addition in RPO is the contextual reweighting factor $\omega_{i,j}$, which measures the similarity between each win–lose pair, where each winner is contrasted against all losers in the batch:

\begin{equation}
\omega_{i,j} =
\frac{\exp\left(-\tfrac{1 - \cos(f(o^w_i), f(o^\ell_j))}{\tau}\right)}
{\sum_{j'} \exp\left(-\tfrac{1 - \cos(f(o^w_i), f(o^\ell_{j'}))}{\tau}\right)}
\label{eq:cos}
\end{equation}

where $f(o^*)$ is an encoder extracting image embeddings and $\tau$ a temperature hyperparameter. This soft alignment focuses the model on semantically similar pairs, ensuring each winner distributes its weight across all losers while giving greater emphasis to losers that are close in semantic space~\cite{yin2024relative}.

\subsubsection{Diffusion-KTO}

Diffusion-KTO~\cite{li2024aligning}, inspired by Kahneman-Tversky Optimization (KTO), aligns diffusion models using only per-sample binary (\textit{win} or \textit{lose}) feedback, rather than paired comparisons. Unlike DPO and RPO, which require preference pairs, KTO incorporates the relative value of each sample with respect to the reference policy, enabling training even with batches containing only winners or losers.

Each sample in $D_{\text{pref}}$ is labeled as winner ($q=1$) or loser ($q=-1$) and at each sampling step $t \sim \mathcal{U}(0,T)$ the deviation of the current policy from the reference policy is computed through:
\begin{multline}
\mathcal{L}_{\text{Diffusion-KTO}}(\theta) =
- \mathbb{E}_{A_0 \sim \mathcal{D}_{\text{pref}}, \, 
t \sim \mathcal{U}(0, T), \,
\epsilon \sim \mathcal{N}(0, I)} \\
\sigma \left(
\beta \cdot q \cdot \left[
\| \epsilon - \epsilon_\theta(O_t, A_t, t) \|_2^2
- \| \epsilon - \epsilon_{\text{ref}}(O_t, A_t, t) \|_2^2
\right] - Q_{\text{ref}}
\right)
\end{multline}
where $\epsilon$ and  $\epsilon_{\text{ref}}$ are as in DPO, $\sigma(\cdot)$ is a sigmoid utility function and $\beta$ controls the scale of the reward signal as in DPO and RPO. 
The sigmoid utility function acts as a smooth preference indicator for each sample, while $Q_{\text{ref}}$ stabilizes training by centering the reward and is constrained to be non-negative as it approximates a KL-divergence term (see Eq. 7 in~\cite{li2024aligning}).

\subsection{Diffusion-RKO}  
We propose Diffusion-RKO, which unifies the benefits of Diffusion-KTO and Diffusion-RPO: like KTO, Diffusion-RKO uses binary preference labels $q \in \{+1, -1\}$ and does not require explicit win–lose pairs; at the same time, it integrates RPO’s similarity-based batch reweighting to emphasize hard negatives and ambiguous winners.
Given a mini-batch of $B$ preference-labeled samples $\{(O_0^b, A_0^b, q)\}_{b=1}^{B}$, $\epsilon_\theta$ and $\epsilon_{\text{ref}}$ are evaluated on noised inputs $(O_t^b, A_t^b)$, with timestep $t \sim \mathcal{U}(0, T)$ and noise $\epsilon \sim \mathcal{N}(0, I)$. The per-sample reward advantage is:
\[
A_b = \beta \left[
\| \epsilon - \epsilon_\theta(O_t^b, A_t^b, t) \|_2^2
- \| \epsilon - \epsilon_{\text{ref}}(O_t^b, A_t^b, t) \|_2^2
\right] - Q_{\text{ref}}
\]

The utility of each sample is computed using a sigmoid function:
\[
U_b = \sigma(q_b \cdot A_b)
\]
 
To amplify the training signal around semantically ambiguous regions, RKO applies a contextual weight $s_b$ to each sample, based on similarity between winners and losers in the batch. Following RPO, a similarity matrix $\omega_{i,j}$ is computed between winner and loser embeddings following Equation~\ref{eq:cos}.

These weights are then aggregated into per-sample scalars:
\begin{align*}
s_i^{\text{pos}} &= 1 + \max_j \omega_{i,j} \qquad &\text{(for each winner } i\text{)} \\
s_j^{\text{neg}} &= \sum_i \omega_{i,j} \qquad &\text{(for each loser } j\text{)}
\end{align*}

After normalization to mean 1, we define $s_b$ for each sample $b$ depending on whether it is a winner or a loser.

The weighted KTO loss is then given by:
\[
\mathcal{L}_{\text{Diffusion-RKO}}(\theta)
= - \frac{1}{\sum_{b=1}^{B} s_b}
\sum_{b=1}^{B} s_b \cdot \sigma\left(q_b \cdot A_b \right)
\]

This formulation allows Diffusion-RKO to focus learning on difficult samples where winners resemble losers and viceversa.

\paragraph{Similarity reweighting and convergence intuition} The similarity matrix $\omega_{i,j}$ defines a kernel on the representation space induced by the encoder embeddings (Eq.~\ref{eq:cos}). Under this view, $s_i^{\text{pos}}$ increases when a preferred sample $i$ has a nearby dispreferred neighbor (large $\max_j \omega_{i,j}$), i.e., when the sample lies close to the local decision boundary between preferred and dispreferred behaviors. Conversely, $s_j^{\text{neg}}$ increases with the local density of preferred neighbors around a dispreferred sample (large $\sum_i \omega_{i,j}$), highlighting \emph{hard negatives} that are difficult to separate in representation space. This yields a principled mechanism: Diffusion-RKO allocates larger gradient mass to regions where preferred and dispreferred behaviors overlap, while leaving isolated, high-margin samples close to the unweighted KTO regime. From an optimization perspective, recent work proves contraction-style stability/convergence guarantees under a softmax bandit abstraction for DPO \cite{shi2024crucial}; since KTO can be cast in the same KL/log-ratio form and RKO applies bounded similarity reweighting to the same per-sample terms, we expect analogous theoretical convergence guarantees to hold for KTO and RKO up to constant rescaling. We refer the reader to our website for the full convergence analysis\footnote{\href{https://github.com/Preference-DOM}{github.com/Preference-DOM}}.


\subsection{Dataset Generation}

For each preference, a policy is trained on 60 winning and 60 losing demonstrations, named $D_{pref}$, with 40 demos shared across all policies and 20 collected via human takeover~\cite{ingelhag2025real}. 
These 20 demonstrations are collected by running the corresponding policy individually, after it has been initially trained on 40 demonstrations. This yields 60 winning demonstrations per preference dataset in total, with takeover demonstrations comprising $50\%$ of the data. Takeover is particularly advantageous as it allows to efficiently collect targeted demonstrations in out-of-distribution states that could be missing from the initial dataset.
Reference policies are trained on a dataset of 100 demonstrations (70 standard, 30 takeover), which we name $D_{ref}$. Each demonstration consists of image observations, actions, and robot state, with an average length of $24s$ for trousers, $29s$ for sleeves and $25s$ for tshirt.
We construct $D_{pref}$ and $D_{ref}$ by rotating the "roles" of three preference datasets in a fixed cycle.
As shown in Fig.~\ref{fig:tasks}, there are three preference sets: pref\_1, pref\_2, and pref\_3. For each preference setting, one set provides \textit{winning} demonstrations, a different set provides \textit{losing} demonstrations, and the remaining set is used to train the reference model (using 100 demonstrations).
The assignment is as follows:
\begin{itemize}
    \item For pref\_1: winners come from pref\_1, losers come from pref\_2, and the reference model is trained on pref\_3.
    \item For pref\_2: winners come from pref\_2, losers come from pref\_3, and the reference model is trained on pref\_1.
    \item For pref\_3: winners come from pref\_3, losers come from pref\_1, and the reference model is trained on pref\_2.
\end{itemize}

This cyclic setup ensures that, for each preference, the \textit{winning} demonstrations match the target preference, while the \textit{losing} demonstrations and the reference model are drawn from distinct, non-overlapping preference sources.

\begin{figure}[t] 
    \centering 
    \includegraphics[width=0.90\linewidth]{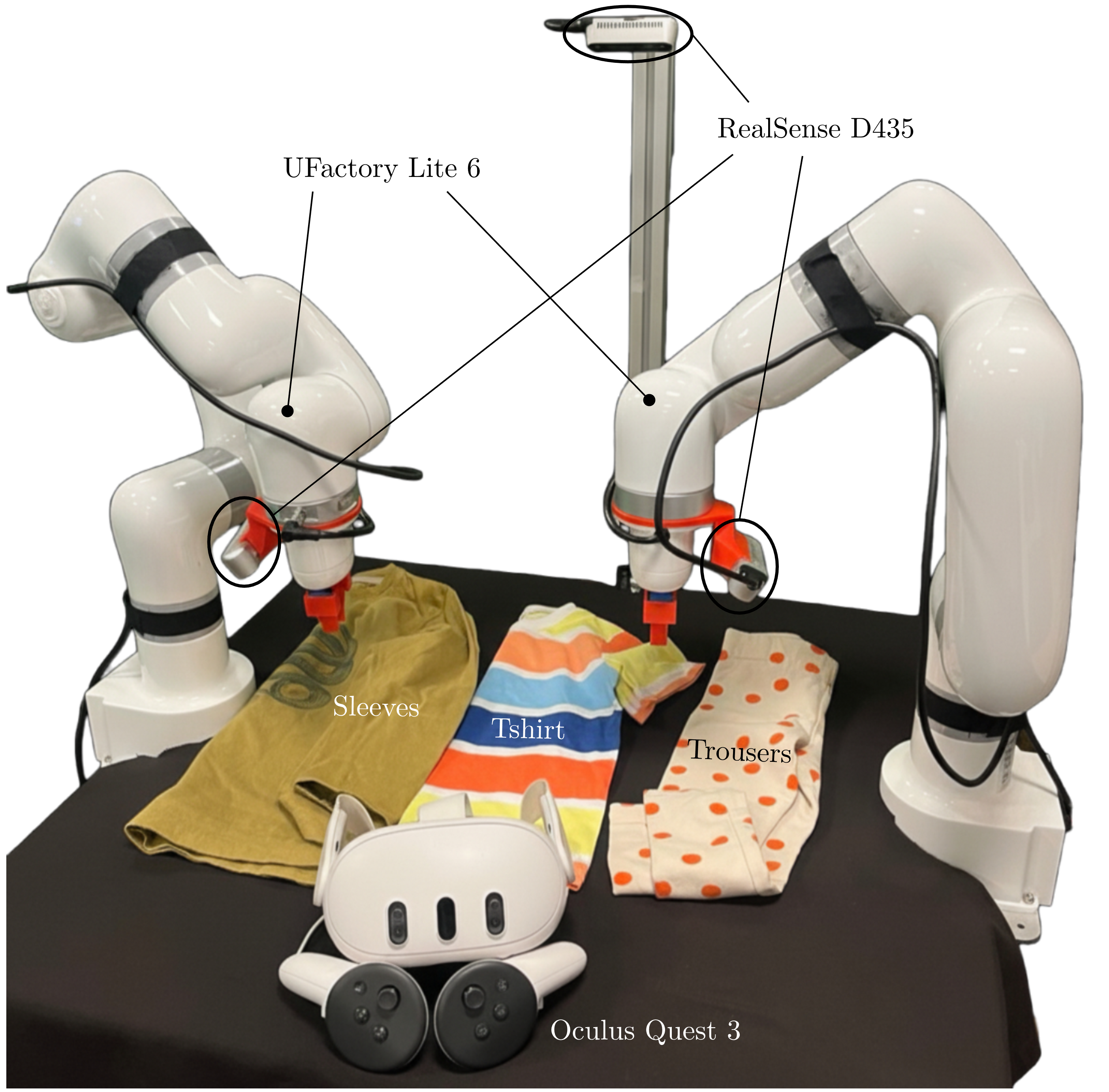} \caption{Experimental setup. } 
    \label{fig:robot_setup} 
\end{figure}

\section{Experiments}
We now describe how we evaluated different preference alignment frameworks. The main goal is to compare the proposed Diffusion-RKO with Diffusion-DPO, -RPO, -KTO, and a vanilla DDPM baseline in terms of sample efficiency. To ensure fairness, all policies are trained on the same set of \textit{winning} demonstrations and evaluated on their ability to reproduce the preferred behavior. We focus on real-world cloth folding, a domain where user preferences strongly influence task execution but are often overlooked. All experiments are conducted on real robots, considering that physically accurate simulation of deformable objects remains challenging~\cite{yin2021modeling} and large-scale demonstration datasets for training diffusion policies on cloth manipulation tasks are scarce both in simulation and the real world. Since our goal is to evaluate the models in terms of sample efficiency, we consider two experimental settings:

\begin{itemize}
\item \textbf{Performance comparison:} we compare all models on a fixed set of $60$ \textit{winning} demonstrations for challenging cloth-folding tasks. We chose $60$ as we observed a good trade-off between policy performance and demonstration collection effort.
\item \textbf{Sample efficiency vs. \# demonstrations:} we compare all models while varying the number of training demonstrations ($20$ to $95$ \textit{winning} demonstrations, in increments of $15$), tracking performance to analyze how each framework scales with the availability of demonstrations. In addition, we perform an ablation experiment for RKO where we set the similarity-reweighting factor to $1$, in order to empirically evaluate its influence.
\end{itemize}

\subsection{Experimental settings and implementation details}
Cloth folding is notoriously difficult to evaluate~\cite{garcia2020benchmarking}. To address this, we adopt a step-wise evaluation scheme based on the correctness of individual pick-and-place actions during execution, providing a more informative signal than binary success/failure. Scoring details for each garment are shown in Fig.~\ref{fig:tasks}, with task scores normalized to sum to 1. A pick action is only counted if it includes a successful lift, making it slightly more valuable than a place action, since failed lifts invalidate the attempt. The aim is not perfect folding, but alignment to the preferred behavior under sample-efficiency constraints. Due to the proximity of many joint singularities, the model must be precise to avoid failure. Executions stop when: the task is completed; the arms enter a singularity or collide with the sponge-covered table; or the system reaches an out-of-distribution state in robot poses or garment configuration from which recovery is impossible. Objects are placed with slight variations across trials, and additional variability arises from garment deformability. Recovery is allowed without penalty, as some demonstrations could include it, and we observe some generalization, with policies recovering autonomously from minor failures.

We use two UFactory Lite6 robots equipped with three RealSense D435 cameras (Fig.~\ref{fig:robot_setup}): two wrist-mounted and one bird’s-eye view. For trousers, only the bird’s-eye and right wrist cameras are used; for sleeves and t-shirts, all three are employed. Demonstrations are collected via teleoperation using the Quest2ROS Oculus app~\cite{welle2024quest2ros}.
Our experiments cover three garment types (trousers, tshirt, and sleeves), and three task variations with different levels of task complexity. Trousers involve single-arm manipulation with two-camera input, while sleeves and tshirt require bimanual coordination and synchronized actions. Each policy is evaluated over 10 runs per task, for a total of 1170 executions. All policies are trained using RGB image observations. The model architecture consists of a CNN encoder and a U-Net-based denoising backbone. DDPM baselines are initialized from a model pretrained on a separate folding task for fairness, though we found no significant performance boost from this. All models are trained for 100k training steps. For preference models, we select checkpoints based on the best reward difference between \textit{winning} and \textit{losing} demonstrations. All models use a batch size of 64 and a learning rate of $3 \cdot 10^{-5}$. Following~\cite{wallace2024diffusion}, we incorporate the time-dependent factor $T$ into the scaling constant $\beta$. We use the following hyperparameters, selected based on their stable and consistent performance during preliminary evaluations:

\begin{itemize}
    \item $\beta = 10$ for DPO,
    \item $\beta = 20$ for RPO,
    \item $\beta = 12$ for KTO and RKO,
    \item $\tau = 0.15$ for RPO and RKO.
\end{itemize}

\subsection{Results}
We report results from two evaluation settings: (1) performance comparison across models using a fixed number of demonstrations, and (2) sample efficiency analysis by varying the number of demonstrations used for training. Videos of the executions and further details are available on our website\footnote{\href{https://github.com/Preference-DOM}{github.com/Preference-DOM}}.

\textbf{Performance comparison: } Table~\ref{tab:trousers}, ~\ref{tab:sleeves} and ~\ref{tab:tshirt} summarize the results for the Trousers, Sleeves and Tshirt performance comparison experiments. 

\begin{table}[h!]
\centering
\renewcommand{\arraystretch}{1.5}
\begin{tabular}{|l|c|c|c|}
\hline
\textbf{Model}   & \textbf{Pref 1}           & \textbf{Pref 2}           & \textbf{Pref 3}          \\ \hline \hline
$\pi_{\text{DDPM}}$             & $0.701 \pm 0.377$           & $0.667 \pm 0.381$           & $0.453 \pm 0.416$          \\ \hline \hline
$\pi_{\text{DPO}}$              & $0.680 \pm 0.336$           & $0.780 \pm 0.225$           & $0.590 \pm 0.348$          \\ \hline
$\pi_{\text{RPO}}$              & $0.790 \pm 0.238$           & $\mathbf{0.790 \pm 0.233}$  & $0.570 \pm 0.271$          \\ \hline
$\pi_{\text{KTO}}$              & $0.860 \pm 0.206$           & $0.650 \pm 0.381$           & $0.520 \pm 0.322$          \\ \hline
$\pi_{\text{RKO}}$ (ours)       & $\mathbf{0.910 \pm 0.166}$  & $0.760 \pm 0.334$           & $\mathbf{0.620 \pm 0.367}$ \\ \hline
\end{tabular}
\caption{Trousers - Performance Comparison}
\label{tab:trousers}
\end{table}

\begin{table}[h!]
\centering
\renewcommand{\arraystretch}{1.5}
\begin{tabular}{|l|c|c|c|}
\hline
\textbf{Model}   & \textbf{Pref 1}           & \textbf{Pref 2}           & \textbf{Pref 3}          \\ \hline \hline
$\pi_{\text{DDPM}}$             & $0.510 \pm 0.301$           & $0.563 \pm 0.306$           & $0.626 \pm 0.279$          \\ \hline \hline
$\pi_{\text{DPO}}$              & $0.570 \pm 0.275$           & $0.350 \pm 0.163$           & $0.460 \pm 0.279$          \\ \hline
$\pi_{\text{RPO}}$              & $0.455 \pm 0.314$           & $0.685 \pm 0.238$           & $0.690 \pm 0.156$          \\ \hline
$\pi_{\text{KTO}}$              & $\mathbf{0.730 \pm 0.249}$  & $0.660 \pm 0.297$           & $\mathbf{0.730 \pm 0.314}$ \\ \hline
$\pi_{\text{RKO}}$ (ours)       & $0.715 \pm 0.274$           & $\mathbf{0.695 \pm 0.207}$  & $0.695 \pm 0.199$          \\ \hline
\end{tabular}
\caption{Sleeves - Performance Comparison}
\label{tab:sleeves}
\end{table}

\begin{table}[h!]
\centering
\renewcommand{\arraystretch}{1.5}
\begin{tabular}{|l|c|c|c|}
\hline
\textbf{Model}   & \textbf{Pref 1}           & \textbf{Pref 2}           & \textbf{Pref 3}          \\ \hline \hline
$\pi_{\text{DDPM}}$             & $0.496 \pm 0.333$           & $0.453 \pm 0.436$           & $0.593 \pm 0.405$          \\ \hline \hline
$\pi_{\text{DPO}}$              & $0.515 \pm 0.368$           & $0.520 \pm 0.307$           & $0.520 \pm 0.391$          \\ \hline
$\pi_{\text{RPO}}$              & $0.525 \pm 0.192$           & $\mathbf{0.560 \pm 0.350}$  & $0.440 \pm 0.338$          \\ \hline
$\pi_{\text{KTO}}$              & $0.595 \pm 0.274$           & $0.540 \pm 0.389$           & $\mathbf{0.695 \pm 0.219}$ \\ \hline
$\pi_{\text{RKO}}$ (ours)       & $\mathbf{0.605 \pm 0.306}$  & $0.550 \pm 0.251$           & $0.660 \pm 0.302$          \\ \hline
\end{tabular}
\caption{Tshirt - Performance Comparison}
\label{tab:tshirt}
\end{table}

The results show that preference optimization frameworks consistently outperform the vanilla DDPM baseline across all tasks. Among individual models, $\pi_{\text{RKO}}$ achieves the best performance in 4 out of 9 tasks, followed by $\pi_{\text{KTO}}$ in 3 and $\pi_{\text{RPO}}$ in 2, suggesting that these methods are generally more effective than DPO for preference alignment and sample efficiency. This highlights the value of using more expressive preference frameworks even in scenarios not directly related to preference alignment but that still use preference optimization frameworks, as in \cite{chen2024deformpam}.
Three of four preference-based methods consistently outperform the DDPM baseline, and RKO does so across all tasks. They are also more training-efficient, typically reaching strong performance within 40–50k steps versus 100k for DDPM. Overall, these results indicate that preference learning (especially RKO, KTO, and RPO) provides substantially better sample and training efficiency than standard DDPM or DPO for aligning diffusion policies to user-preferred behaviors.

\begin{figure}[t] 
    \centering 
    \includegraphics[width=0.99\linewidth]{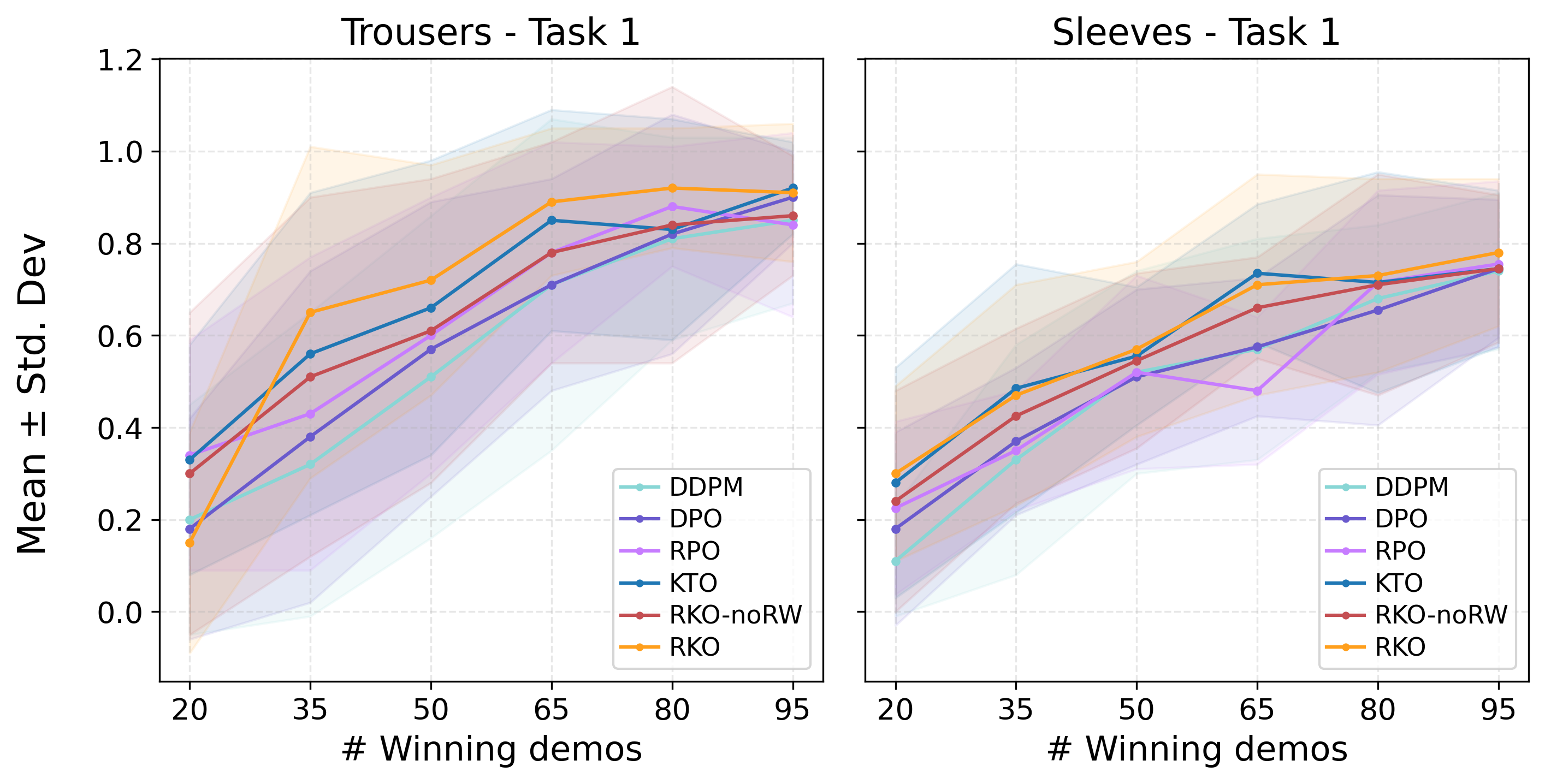} \caption{Sample efficiency experiment: influence of the number of demonstrations ($20$ to $95$ \textit{winning} demos, increments of $15$) on the performance, in Trousers - Pref 1 and Sleeves - Pref 1.} 
    \label{fig:exp2} 
\end{figure}

\textbf{Sample efficiency vs. \# demos: } results are shown in Fig.~\ref{fig:exp2}: RKO consistently outperforms DDPM in terms of sample efficiency across all settingsand typically matches or outperforms the other baselines. The main exceptions are Sleeves Task~1, where KTO performs better at 35 and 65 demonstrations, and Trousers Task~1 with 20 demonstrations, where RKO performs worst. In the latter low-data regime, we frequently observe the robot going out-of-distribution during executions, likely due to the limited number of \textit{winning} and \textit{losing} demonstrations which may confuse RKO during training. This effect vanishes as the dataset grows, after which RKO again surpasses DDPM, indicating that its advantage is robust when sufficient preference data are available. For the ablated variant without reweighting (RKO-noRW), performance is consistently lower than full RKO across the experiments, indicating that the reweighting term contributes in boosting RKO.
Given the relatively large standard deviations observed in the experiments, we further assessed the robustness of the results using a Bayesian A/B test, following the approach discussed in~\cite{kress2024robot}, defining a task as successful when the achieved score is at least 0.75. The posterior analysis shows that RKO outperforms DDPM with probabilities of $87\%$ on Trousers and $70.1\%$ on Sleeves, consistent across varying numbers of demonstrations, providing statistical evidence of RKO’s reliability over DDPM. 

Overall, results from both experiments indicate that using a preference alignment framework is generally more effective than training a vanilla diffusion model from scratch, especially when demonstrations of alternative behaviors and pretrained reference policies are available. Preference-aligned policies not only learn the desired behavior more accurately, but also benefit from being explicitly constrained to avoid dispreferred behaviors. This leads to improved performance and generally faster training. These advantages are particularly evident in scenarios where additional demonstrations (even if not representing the preferred behavior) can be repurposed as \textit{losing} examples instead of being discarded.
In terms of failure modes, we observe no method-specific failure patterns across the different preference optimization frameworks. Most failures arise when the robot enters states that are out of distribution relative to the training data, often due to imprecise end-effector orientation or incorrect initial actions that lead to unseen states. Failures are more frequent in bimanual tasks, where coordination between arms increases the likelihood of distributional shift, and are more pronounced for DDPM- and DPO-based policies.

\section{Conclusion}

In this work, we studied alignment of pretrained visuomotor diffusion policies to user-preferred behaviors for deformable object manipulation, and provided the first systematic comparison of diffusion preference objectives (DPO, RPO, KTO) while introducing RKO, which combines the benefits of KTO and RPO. Across multiple real-world cloth-folding tasks, preference alignment consistently improves performance and sample efficiency over standard DDPM fine-tuning, suggesting a practical path to incorporating user preferences using limited demonstrations instead of retraining from scratch. Future work will study robustness to noisy or suboptimal demonstrations, automatic selection of negative samples from offline data, and extension beyond cloth folding to other types of deformable objects.


\balance
\bibliographystyle{ieeetr}
\bibliography{references}

@inproceedings{zhang2017personalized,
  title={Personalized robot-assisted dressing using user modeling in latent spaces},
  author={Zhang, Fan and Cully, Antoine and Demiris, Yiannis},
  booktitle={2017 IEEE/RSJ International Conference on Intelligent Robots and Systems (IROS)},
  pages={3603--3610},
  year={2017},
  organization={IEEE}
}

@article{ho2020denoising,
  title={Denoising diffusion probabilistic models},
  author={Ho, Jonathan and Jain, Ajay and Abbeel, Pieter},
  journal={Advances in neural information processing systems},
  volume={33},
  pages={6840--6851},
  year={2020}
}

@article{li2024aligning,
  title={Aligning diffusion models by optimizing human utility},
  author={Li, Shufan and Kallidromitis, Konstantinos and Gokul, Akash and Kato, Yusuke and Kozuka, Kazuki},
  journal={Advances in Neural Information Processing Systems},
  volume={37},
  pages={24897--24925},
  year={2024}
}

@article{yin2024relative,
  title={Relative preference optimization: Enhancing llm alignment through contrasting responses across identical and diverse prompts},
  author={Yin, Yueqin and Wang, Zhendong and Gu, Yi and Huang, Hai and Chen, Weizhu and Zhou, Mingyuan},
  journal={arXiv preprint arXiv:2402.10958},
  year={2024}
}

@inproceedings{sridhar2024nomad,
  title={Nomad: Goal masked diffusion policies for navigation and exploration},
  author={Sridhar, Ajay and Shah, Dhruv and Glossop, Catherine and Levine, Sergey},
  booktitle={2024 IEEE International Conference on Robotics and Automation (ICRA)},
  pages={63--70},
  year={2024},
  organization={IEEE}
}

@article{pearce2023imitating,
  title={Imitating human behaviour with diffusion models},
  author={Pearce, Tim and Rashid, Tabish and Kanervisto, Anssi and Bignell, Dave and Sun, Mingfei and Georgescu, Raluca and Macua, Sergio Valcarcel and Tan, Shan Zheng and Momennejad, Ida and Hofmann, Katja and others},
  journal={arXiv preprint arXiv:2301.10677},
  year={2023}
}

@article{chi2023diffusion,
  title={Diffusion policy: Visuomotor policy learning via action diffusion},
  author={Chi, Cheng and Xu, Zhenjia and Feng, Siyuan and Cousineau, Eric and Du, Yilun and Burchfiel, Benjamin and Tedrake, Russ and Song, Shuran},
  journal={The International Journal of Robotics Research},
  pages={02783649241273668},
  year={2023},
  publisher={SAGE Publications Sage UK: London, England}
}

@article{reuss2023goal,
  title={Goal-conditioned imitation learning using score-based diffusion policies},
  author={Reuss, Moritz and Li, Maximilian and Jia, Xiaogang and Lioutikov, Rudolf},
  journal={arXiv preprint arXiv:2304.02532},
  year={2023}
}

@ARTICLE{kapelyukh2023dallebot,
  author={Kapelyukh, Ivan and Vosylius, Vitalis and Johns, Edward},
  journal={IEEE Robotics and Automation Letters}, 
  title={DALL-E-Bot: Introducing Web-Scale Diffusion Models to Robotics}, 
  year={2023},
  volume={8},
  number={7},
  pages={3956-3963},
  doi={10.1109/LRA.2023.3272516}}

@InProceedings{mishra2023generative,
  title = 	 {Generative Skill Chaining: Long-Horizon Skill Planning with Diffusion Models},
  author =       {Mishra, Utkarsh Aashu and Xue, Shangjie and Chen, Yongxin and Xu, Danfei},
  booktitle = 	 {Proceedings of The 7th Conference on Robot Learning},
  pages = 	 {2905--2925},
  year = 	 {2023},
  editor = 	 {Tan, Jie and Toussaint, Marc and Darvish, Kourosh},
  volume = 	 {229},
  series = 	 {Proceedings of Machine Learning Research},
  month = 	 {06--09 Nov},
  publisher =    {PMLR},
  url = 	 {https://proceedings.mlr.press/v229/mishra23a.html}
}

@inproceedings{cheng2011preference,
  title={Preference-based policy iteration: Leveraging preference learning for reinforcement learning},
  author={Cheng, Weiwei and F{\"u}rnkranz, Johannes and H{\"u}llermeier, Eyke and Park, Sang-Hyeun},
  booktitle={Joint European Conference on Machine Learning and Knowledge Discovery in Databases},
  pages={312--327},
  year={2011},
  organization={Springer}
}

@article{ouyang2022training,
  title={Training language models to follow instructions with human feedback},
  author={Ouyang, Long and Wu, Jeffrey and Jiang, Xu and Almeida, Diogo and Wainwright, Carroll and Mishkin, Pamela and Zhang, Chong and Agarwal, Sandhini and Slama, Katarina and Ray, Alex and others},
  journal={Advances in neural information processing systems},
  volume={35},
  pages={27730--27744},
  year={2022}
}

@article{wilson2012bayesian,
  title={A bayesian approach for policy learning from trajectory preference queries},
  author={Wilson, Aaron and Fern, Alan and Tadepalli, Prasad},
  journal={Advances in neural information processing systems},
  volume={25},
  year={2012}
}

@article{yuan2024preference,
  title={Preference aligned diffusion planner for quadrupedal locomotion control},
  author={Yuan, Xinyi and Shang, Zhiwei and Wang, Zifan and Wang, Chenkai and Shan, Zhao and Zhu, Meixin and Bai, Chenjia and Li, Xuelong and Wan, Weiwei and Harada, Kensuke},
  journal={arXiv preprint arXiv:2410.13586},
  year={2024}
}

@inproceedings{lee2024visual,
  title={Visual Preference Inference: An Image Sequence-Based Preference Reasoning in Tabletop Object Manipulation},
  author={Lee, Joonhyung and Park, Sangbeom and Kwon, Yongin and Lee, Jemin and Ahn, Minwook and Choi, Sungjoon},
  booktitle={2024 IEEE/RSJ International Conference on Intelligent Robots and Systems (IROS)},
  pages={9745--9752},
  year={2024},
  organization={IEEE}
}

@article{canal2019adapting,
  title={Adapting robot task planning to user preferences: an assistive shoe dressing example},
  author={Canal, Gerard and Aleny{\`a}, Guillem and Torras, Carme},
  journal={Autonomous Robots},
  volume={43},
  number={6},
  pages={1343--1356},
  year={2019},
  publisher={Springer}
}

@article{winata2025preference,
  title={Preference tuning with human feedback on language, speech, and vision tasks: A survey},
  author={Winata, Genta Indra and Zhao, Hanyang and Das, Anirban and Tang, Wenpin and Yao, David D and Zhang, Shi-Xiong and Sahu, Sambit},
  journal={Journal of Artificial Intelligence Research},
  volume={82},
  pages={2595--2661},
  year={2025}
}

@article{rafailov2023direct,
  title={Direct preference optimization: Your language model is secretly a reward model},
  author={Rafailov, Rafael and Sharma, Archit and Mitchell, Eric and Manning, Christopher D and Ermon, Stefano and Finn, Chelsea},
  journal={Advances in neural information processing systems},
  volume={36},
  pages={53728--53741},
  year={2023}
}

@article{palan2019learning,
  title={Learning reward functions by integrating human demonstrations and preferences},
  author={Palan, Malayandi and Landolfi, Nicholas C and Shevchuk, Gleb and Sadigh, Dorsa},
  journal={arXiv preprint arXiv:1906.08928},
  year={2019}
}

@article{ingelhag2025real,
  title={Real-Time Operator Takeover for Visuomotor Diffusion Policy Training},
  author={Ingelhag, Nils and Munkeby, Jesper and Welle, Michael C and Moletta, Marco and Kragic, Danica},
  journal={arXiv preprint arXiv:2502.02308},
  year={2025}
}

@inproceedings{ingelhag2024robotic,
  title={A robotic skill learning system built upon diffusion policies and foundation models},
  author={Ingelhag, Nils and Munkeby, Jesper and van Haastregt, Jonne and Varava, Anastasia and Welle, Michael C and Kragic, Danica},
  booktitle={2024 33rd IEEE International Conference on Robot and Human Interactive Communication (ROMAN)},
  pages={748--754},
  year={2024},
  organization={IEEE}
}

@article{dalal2023imitating,
  title={Imitating task and motion planning with visuomotor transformers},
  author={Dalal, Murtaza and Mandlekar, Ajay and Garrett, Caelan and Handa, Ankur and Salakhutdinov, Ruslan and Fox, Dieter},
  journal={arXiv preprint arXiv:2305.16309},
  year={2023}
}

@misc{black2024pi0,
      title={$\pi_0$: A Vision-Language-Action Flow Model for General Robot Control}, 
      author={Kevin Black and Noah Brown and Danny Driess and Adnan Esmail and Michael Equi and Chelsea Finn and Niccolo Fusai and Lachy Groom and Karol Hausman and Brian Ichter and Szymon Jakubczak and Tim Jones and Liyiming Ke and Sergey Levine and Adrian Li-Bell and Mohith Mothukuri and Suraj Nair and Karl Pertsch and Lucy Xiaoyang Shi and James Tanner and Quan Vuong and Anna Walling and Haohuan Wang and Ury Zhilinsky},
      year={2024},
      eprint={2410.24164},
      archivePrefix={arXiv},
      primaryClass={cs.LG},
      url={https://arxiv.org/abs/2410.24164}, 
}

@incollection{furnkranz2010preference,
  title={Preference learning and ranking by pairwise comparison},
  author={F{\"u}rnkranz, Johannes and H{\"u}llermeier, Eyke},
  booktitle={Preference learning},
  pages={65--82},
  year={2010},
  publisher={Springer}
}

@book{sadigh2017active,
  title={Active preference-based learning of reward functions},
  author={Sadigh, Dorsa and Dragan, Anca and Sastry, Shankar and Seshia, Sanjit},
  year={2017}
}

@inproceedings{woodworth2018preference,
  title={Preference learning in assistive robotics: Observational repeated inverse reinforcement learning},
  author={Woodworth, Bryce and Ferrari, Francesco and Zosa, Teofilo E and Riek, Laurel D},
  booktitle={Machine learning for healthcare conference},
  pages={420--439},
  year={2018},
  organization={PMLR}
}

@article{wirth2017survey,
  title={A survey of preference-based reinforcement learning methods},
  author={Wirth, Christian and Akrour, Riad and Neumann, Gerhard and F{\"u}rnkranz, Johannes},
  journal={Journal of Machine Learning Research},
  volume={18},
  number={136},
  pages={1--46},
  year={2017}
}

@article{jain2015learning,
  title={Learning preferences for manipulation tasks from online coactive feedback},
  author={Jain, Ashesh and Sharma, Shikhar and Joachims, Thorsten and Saxena, Ashutosh},
  journal={The International Journal of Robotics Research},
  volume={34},
  number={10},
  pages={1296--1313},
  year={2015},
  publisher={SAGE Publications Sage UK: London, England}
}

@inproceedings{cakmak2011human,
  title={Human preferences for robot-human hand-over configurations},
  author={Cakmak, Maya and Srinivasa, Siddhartha S and Lee, Min Kyung and Forlizzi, Jodi and Kiesler, Sara},
  booktitle={2011 IEEE/RSJ International Conference on Intelligent Robots and Systems},
  pages={1986--1993},
  year={2011},
  organization={IEEE}
}

@article{chen2024deformpam,
  title={DeformPAM: Data-Efficient Learning for Long-horizon Deformable Object Manipulation via Preference-based Action Alignment},
  author={Chen, Wendi and Xue, Han and Zhou, Fangyuan and Fang, Yuan and Lu, Cewu},
  journal={arXiv preprint arXiv:2410.11584},
  year={2024}
}

@inproceedings{wallace2024diffusion,
  title={Diffusion model alignment using direct preference optimization},
  author={Wallace, Bram and Dang, Meihua and Rafailov, Rafael and Zhou, Linqi and Lou, Aaron and Purushwalkam, Senthil and Ermon, Stefano and Xiong, Caiming and Joty, Shafiq and Naik, Nikhil},
  booktitle={Proceedings of the IEEE/CVF Conference on Computer Vision and Pattern Recognition},
  pages={8228--8238},
  year={2024}
}

@article{gu2024diffusion,
  title={Diffusion-rpo: Aligning diffusion models through relative preference optimization},
  author={Gu, Yi and Wang, Zhendong and Yin, Yueqin and Xie, Yujia and Zhou, Mingyuan},
  journal={arXiv preprint arXiv:2406.06382},
  year={2024}
}

@article{longhini2024unfolding,
  title={Unfolding the literature: A review of robotic cloth manipulation},
  author={Longhini, Alberta and Wang, Yufei and Garcia-Camacho, Irene and Blanco-Mulero, David and Moletta, Marco and Welle, Michael and Aleny{\`a}, Guillem and Yin, Hang and Erickson, Zackory and Held, David and others},
  journal={Annual Review of Control, Robotics, and Autonomous Systems},
  volume={8},
  year={2024},
  publisher={Annual Reviews}
}

@inproceedings{moletta2023virtual,
  title={A virtual reality framework for human-robot collaboration in cloth folding},
  author={Moletta, Marco and Wozniak, Maciej K and Welle, Michael C and Kragic, Danica},
  booktitle={2023 IEEE-RAS 22nd International Conference on Humanoid Robots (Humanoids)},
  pages={1--7},
  year={2023},
  organization={IEEE}
}

@article{laezza2024robot,
  title={Robot Learning for Deformable Object Manipulation Tasks},
  author={Laezza, Rita},
  year={2024}
}

@phdthesis{longhini2025adapting,
  title={Adapting to Variations in Textile Properties for Robotic Manipulation},
  author={Longhini, Alberta},
  year={2025},
  school={KTH Royal Institute of Technology}
}

@article{yin2021modeling,
  title={Modeling, learning, perception, and control methods for deformable object manipulation},
  author={Yin, Hang and Varava, Anastasia and Kragic, Danica},
  journal={Science Robotics},
  volume={6},
  number={54},
  pages={eabd8803},
  year={2021},
  publisher={American Association for the Advancement of Science}
}

@article{canal2020adapting,
  title={Adapting robot behavior to user preferences in assistive scenarios},
  author={Canal, Gerard},
  year={2020},
  publisher={Universidad Polit{\'e}cnica de Catalu{\~n}a}
}

@inproceedings{welle2024quest2ros,
  title={Quest2ros: An app to facilitate teleoperating robots},
  author={Welle, Michael C and Ingelhag, Nils and Lippi, Martina and Wozniak, Maciej and Gasparri, Andrea and Kragic, Danica},
  booktitle={7th International Workshop on Virtual, Augmented, and Mixed-Reality for Human-Robot Interactions},
  year={2024}
}

@article{zhu2022challenges,
  title={Challenges and outlook in robotic manipulation of deformable objects},
  author={Zhu, Jihong and Cherubini, Andrea and Dune, Claire and Navarro-Alarcon, David and Alambeigi, Farshid and Berenson, Dmitry and Ficuciello, Fanny and Harada, Kensuke and Kober, Jens and Li, Xiang and others},
  journal={IEEE Robotics \& Automation Magazine},
  volume={29},
  number={3},
  pages={67--77},
  year={2022},
  publisher={IEEE}
}

@article{liu2025survey,
  title={A survey of direct preference optimization},
  author={Liu, Shunyu and Fang, Wenkai and Hu, Zetian and Zhang, Junjie and Zhou, Yang and Zhang, Kongcheng and Tu, Rongcheng and Lin, Ting-En and Huang, Fei and Song, Mingli and others},
  journal={arXiv preprint arXiv:2503.11701},
  year={2025}
}

@article{garcia2020benchmarking,
  title={Benchmarking bimanual cloth manipulation},
  author={Garcia-Camacho, Irene and Lippi, Martina and Welle, Michael C and Yin, Hang and Antonova, Rika and Varava, Anastasiia and Borras, Julia and Torras, Carme and Marino, Alessandro and Alenya, Guillem and others},
  journal={IEEE Robotics and Automation Letters},
  volume={5},
  number={2},
  pages={1111--1118},
  year={2020},
  publisher={IEEE}
}

@article{kress2024robot,
  title={Robot learning as an empirical science: Best practices for policy evaluation},
  author={Kress-Gazit, Hadas and Hashimoto, Kunimatsu and Kuppuswamy, Naveen and Shah, Paarth and Horgan, Phoebe and Richardson, Gordon and Feng, Siyuan and Burchfiel, Benjamin},
  journal={arXiv preprint arXiv:2409.09491},
  year={2024}
}

@article{shi2024crucial,
  title={The crucial role of samplers in online direct preference optimization},
  author={Shi, Ruizhe and Zhou, Runlong and Du, Simon S},
  journal={arXiv preprint arXiv:2409.19605},
  year={2024}
}


\end{document}